\documentclass[runningheads]{llncs}

\usepackage{eccv}

\usepackage{graphicx}
\usepackage{booktabs}

\usepackage[accsupp]{axessibility}  

\usepackage{hyperref}

\usepackage{orcidlink}

\usepackage{listings}
\usepackage{xcolor}
\definecolor{codegreen}{rgb}{0,0.6,0}
\definecolor{codegray}{rgb}{0.5,0.5,0.5}
\definecolor{codepurple}{rgb}{0.58,0,0.82}
\definecolor{backcolour}{rgb}{0.95,0.95,0.92}

\usepackage{multirow}
\usepackage{tabularx}
\usepackage{ragged2e} 
\newcolumntype{Y}{>{\Centering\arraybackslash}X}
\newcolumntype{M}[1]{>{\Centering\arraybackslash}m{#1}}

\lstdefinestyle{mystyle}{
    language=Python,
    backgroundcolor=\color{backcolour},   
    commentstyle=\color{codegreen},
    keywordstyle=\color{magenta},
    numberstyle=\tiny\color{codegray},
    stringstyle=\color{codepurple},
    basicstyle=\tt\small,  
    breakatwhitespace=false,         
    breaklines=true,                 
    captionpos=b,                    
    keepspaces=true,                 
    numbers=left,                    
    numbersep=5pt,                  
    showspaces=false,                
    showstringspaces=false,
    showtabs=false,                  
    tabsize=4
}
\usepackage{lipsum}
\newcommand*{\mybox}[2]{\colorbox{#1!30}{\parbox{\linewidth}{\fontsize{8}{10} \noindent \newline#2}}}

\begin{document}

\title{HawkEye: Training Video-Text LLMs for Grounding Text in Videos} 



\author{Yueqian Wang\inst{1} \and
Xiaojun Meng\inst{2} \and
Jianxin Liang\inst{1} \and
Yuxuan Wang\inst{3} \and
Qun Liu\inst{2} \and
Dongyan Zhao\inst{1,4} \thanks{Corresponding author}
}

\authorrunning{Y. Wang et al.}


\institute{
Wangxuan Institute of Computer Technology, Peking University \\
\email{\{wangyueqian, liangjx, zhaodongyan\}@pku.edu.cn} \and
Huawei Noah’s Ark Lab \\
\email{\{xiaojun.meng, qun.liu\}@huawei.com} \and
Beijing Institute for General Artificial Intelligence \\
\email{\{wangyuxuan1\}@bigai.ai} \and
National Key Laboratory of General Artificial Intelligence \\
}

\maketitle

\begin{abstract}
Video-text Large Language Models (video-text LLMs) have shown remarkable performance in answering questions and holding conversations on simple videos. 
However, they perform almost the same as random on grounding text queries in long and complicated videos, having little ability to understand and reason about temporal information, which is the most fundamental difference between videos and images. 
In this paper, we propose HawkEye, one of the first video-text LLMs that can perform temporal video grounding in a fully text-to-text manner. To collect training data that is applicable for temporal video grounding, we construct InternVid-G, a large-scale video-text corpus with segment-level captions and negative spans, with which we introduce two new time-aware training objectives to video-text LLMs. We also propose a coarse-grained method of representing segments in videos, which is more robust and easier for LLMs to learn and follow than other alternatives. Extensive experiments show that HawkEye is better at temporal video grounding and comparable on other video-text tasks with existing video-text LLMs, which verifies its superior video-text multi-modal understanding abilities.

\keywords{Video Understanding \and Temporal Video Grounding \and Multi-modal Pre-training}
\end{abstract}

\section{Introduction} \label{sec:intro}
Video-text large language models (LLMs) have been developed rapidly in recent years to help people process videos easier and faster. This development process includes the emergence of a number of new training corpora and models. 
However, most of the training corpora only include short videos with simple contents, in which a single key frame often retains almost all the semantic information of the entire video. As a result, though models trained on these corpora can hold conversations and answer questions regarding short and simple videos, they do little to help us understand long-form videos like movies, tutorials, and documentaries that play an integral role in our daily lives and convey a wealth of information, knowledge, opinions, and emotions. In fact, understanding long-form videos can be very difficult for computers: they first have to understand the basic content and then the sequence of occurrence of multiple events that appear in the video.

To tackle with this problem, in this paper we investigate solving temporal video grounding, a basic task for long-form video understanding, with video-text LLMs. The goal of temporal video grounding is to find a specific segment related to a given text query from a long-form video which contains many actions and events. 
However, existing LLMs perform far from satisfactory on this task. For example, MVBench \cite{Li2023MVBenchAC} and VITATECS \cite{Li2023VITATECSAD} point out that even state-of-the-art video-text LLMs perform like chance on localizing actions or determining the order of events in videos, showing that though being the most substantial difference between videos and images, understanding temporal information in videos have still been ignored by most video-text LLMs.

In this paper, we aim to enhance the temporal video grounding abilities of existing video-text LLMs with targeted training on long-form videos. Due to the computing expenses, an ideal solution is not to modify and rerun the heavy pre-training or visual-text alignment stages, but mainly improve video-text LLMs in the following two aspects: (1) designing good representation formats for a LLM to refer to segments of the video input in text, and (2) constructing a large-scale time-aware instructing training dataset, and jointly train the model on it as well as other instruction datasets.
To improve aspect (1), we try different methods of representing the video segments, and conducted experiments to compare the pros and cons of these methods. We find that rather than directly asking the LLM to generate start and end frame or timestamps of video segments, it's better to refer to video segments in a coarser granularity, \textit{i.e.,} only judging whether a referred segment is in the beginning, the middle, or the end part of the video. With our proposed recursive grounding technique, this representation method can be used to refer to shorter and finer-grained video segments through multiple rounds of judgement. 
To improve aspect (2), we build InternVid-G, a large scale video corpus with 715k segment-level captions and negative spans, which is suitable for constructing temporal video grounding training samples.

Based on the stage 2 checkpoint of VideoChat2 \cite{Li2023MVBenchAC}, by implementing the above improvements we train HawkEye, a video-text LLM with the ability to accomplish temporal video grounding task in a text-to-text manner.
We evaluate its performance on various downstream benchmarks including temporal video grounding, question grounding and video question answering. Experimental results show that HawkEye is one of the first video-text LLMs that performs substantially better than random on temporal video grounding in fully text-to-text manner without hurting the performance on other video-text tasks. We also make several in-depth studies on our propose coarse-grained method of representing video segments to validate its performace and robustness. Code and data is available at: \href{https://github.com/yellow-binary-tree/HawkEye}{https://github.com/yellow-binary-tree/HawkEye}

\section{Related Works}
\subsection{Temporal Video Grounding} \label{sec:related_grounding}
Temporal video grounding, also named as video moment retrieval or video localization, is a task that aims to find segments related to a given text query in videos. Traditional approaches of temporal video grounding can be categorized into two categories: proposal-based approaches \cite{Zhang2019Learning2T,Yuan2019SemanticCD,Zhang2018MANMA,Liu2021AdaptivePG} first propose a set of candidate segments and rank them based on their relevance to the query, while proposal-free approaches \cite{Yuan2018ToFW,Zhang2020SpanbasedLN,Liu2022ReducingTV} directly predict the start and end positions of the segment. 
Charades-STA \cite{Gao2017TALLTA}, ActivityNet-Captions \cite{Krishna2017DenseCaptioningEI}, DiDeMo \cite{Hendricks2017LocalizingMI} and TACoS \cite{Regneri2013GroundingAD} are the commonly-used datasets for training and testing temporal video grounding models. However, these datasets are relatively small in sizes, containing only tens of thousands of videos and queries, due to the expensive cost and labour efforts of annotating training data. To avoid it, weakly-supervised methods \cite{Mithun2019WeaklySV,Yang2021LocalCN,Lin2019WeaklySupervisedVM,Duan2018WeaklySD} that train models without human-annotated ground truth spans are also intensively studied recently. We tackle with this problem from another perspective by automatically annotating a large amount of queries and spans for temporal video grounding training, and finetune existing video-text LLMs in a lightweight text-to-text manner.

\subsection{Video-Text Multimodal LLMs}
With the development of image-text LLMs \cite{Li2023BLIP2BL,Dai2023Instructblip,liu2023llava}, many works aim to combine LLMs with video encoders to leverage the comprehension and generation capabilities of LLMs for video-related tasks \cite{Zhang2023VideoLLaMAAI,Maaz2023VideoChatGPTTD,Lyu2023MacawLLMML,Luo2023ValleyVA,Li2023VideoChatCV,Li2023MVBenchAC}.
These models usually include two phases of training: In stage 1, large-scale video-text paired corpora are used to align the video features to the hidden states of LLMs, and in stage 2, a smaller amount of high-quality LLM-generated or human-annotated conversations are used for instruction tuning. However, as videos in these corpora and datasets are usually short and the conversations often do not focus on temporal information, these models thus often perform poorly in understanding long-form videos or localizing actions or events.

Recently, there have been some attempts to incorporate localization abilities or long-form video understanding into video-text LLMs. SeViLA \cite{Yu2023SelfChainedIM} proposes a localizer to assign a relevance score to each frame in the video, which is then used to filter the relevant frames for video question answering. MovieChat \cite{Song2023MovieChatFD} proposes a memory mechanism to handle with very long video inputs such as movies. 
Kosmos-2 \cite{Peng2023Kosmos2GM} and Qwen-VL \cite{Qwen-VL} shows that LLMs can accomplish a wider variety of downstream tasks if they possess the ability of representing a part of the visual input (\textit{e.g.,} regions of images or segments of videos) in text. Concurrent to our work, VTimeLLM \cite{Huang2023VTimeLLMEL} and TimeChat \cite{Ren2023TimeChatAT} explores to accomplish temporal video grounding in a fully text-to-text manner by collecting instruction tuning datasets for this task. Our work differs from theirs in terms of using different formats to represent video segments, thus achieving better performance in temporal video grounding. In addition, they specifically target video grounding tasks by leveraging LLMs while our motivation is to train a general video-text LLM that still owns versatility on various tasks. Therefore, we also pay efforts on formatting visual grounding similar to other tasks and thus jointly training with many other video-text tasks.

\section{InternVid-G: A Large-Scale Video-Text Dataset with Scene-Level Annotations for Temporal Grounding}

\begin{figure}[t]
  \centering
  \includegraphics[width=0.95\textwidth]{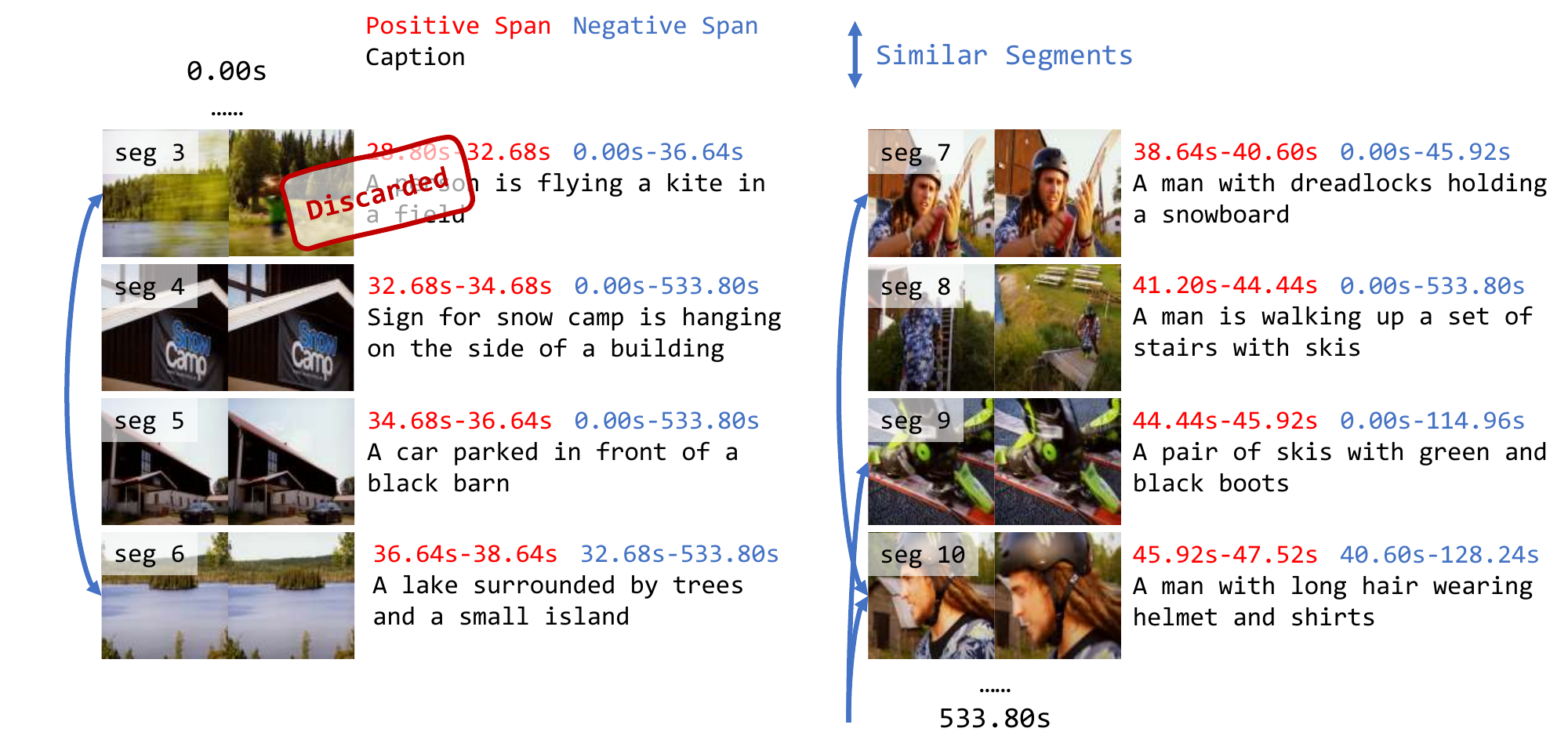}
  \caption{An overview of the InternVid-G dataset. Segment 3 is discarded due to its similarity with the given caption is lower than a threshold. Segment 10's negative span starts from $40.60s$ (in \textcolor{blue}{blue}) as this timestamp is the end position of its similar segment 7, which should not be included since the caption ``\texttt{A man with long hair wearing helmet and shirts}'' of segment 10 also owns a high similarity with Segment 7.}
  \label{fig:dataset}
\end{figure}

\subsection{Dataset Construction}
The pressing matter for improving LLMs' ability on temporal video grounding is to construct a large-scale training dataset for this task. Different from existing large-scale video-text datasets like WebVid \cite{Bain2021FrozenIT} which only contains short videos and corresponding captions, the dataset we plan to use for temporal video grounding training needs to meet the following requirements: (1) The videos should be long and contain multiple events; (2) Captions are annotated at scene level, \textit{i.e.,} each caption should be paired with a segment of the video with a certain start position and end position; (3) Captions need to correspond to the semantic content of video scenes, instead of simply using ASR results of the corresponding audio like HowTo100M \cite{Miech2019HowTo100MLA}.

Therefore, we construct InternVid-G, a dataset that meets all of the above requirements. Fig. \ref{fig:dataset} shows an overview of InternVid-G with 8 video segments. This dataset is constructed through the following steps:

\subsubsection{Scene Segmentation.} 
We randomly download 100k (1\%) videos from InternVid-10M-FLT \cite{Wang2023InternVidAL}, as these videos cover diverse categories and cultural backgrounds. 
We use PySceneDetect\footnote{https://github.com/Breakthrough/PySceneDetect} to split video into scenes. However, PySceneDetect splits the video by detecting abrupt changes in pixels of adjacent frames. In contrast, as we aim to segment videos into scenes with different semantic content, this toolkit results in a number of false positive segmentations (\textit{e.g.,} splitting the video of the same event from different camera angles into different scenes, which is not what we desire). To tackle with this problem, we use CLIP \cite{Radford2021LearningTV} to calculate the semantic similarity between each pair of adjacent scenes, and merge the adjacent scenes if the similarity score between them is higher than a threshold.

\subsubsection{Scene Captioning and Filtering.}
For each video segment, we sample the centre frame and use BLIP-2 \cite{Li2023BLIP2BL} to generate a caption. To ensure the quality of the captions, we use CLIP to calculate the similarity between the caption and the video segment, and only keep half of all the samples that own higher similarity than a threshold. Take Fig \ref{fig:dataset} as an example: the segment 3 owns a relatively lower similarity score with its caption, and is discarded.

\subsubsection{Negative Span Mining.}   
Temporal video grounding requires a model to retrieve a video segment which is relevant to the given text query from a long video. To construct samples for this task, in addition to the video segment and its corresponding caption as query, we also need several other segments before and after this segment as the \textit{video context} to retrieve from.
We term the video segment corresponding to the query as the \textit{positive span}, and video context + positive span as the \textit{negative span}, so models are required to locate the positive span inside the negative span to perform video grounding.

One notable issue is the negative span should not contain other video segments that are too similar to the positive span, otherwise these segments can also correspond to the query and will introduce noises. To prevent this, we calculate the similarity with CLIP between the positive span and all other segments in the video, and label the segments with a similarity score above a threshold as \textit{similar segments}. Thus, for a particular text query paired with its positive span, the start position of its paired negative span can be the end position of the last \textit{similar segment} before its positive span. Similarly, the end position of its paired negative span can be the start position of the first \textit{similar segment} after its positive span. We term these two positions as \texttt{neg\_start} and \texttt{neg\_end}, and the start \& end position of the positive span as \texttt{pos\_start} and \texttt{pos\_end}. If there are no similar segments before or after the positive span, then the \texttt{neg\_start} or \texttt{neg\_end} will be set to 0 (\textit{i.e.,} the beginning of the video) or the end position of the last segment in the video. Examples of negative span mining are also shown in Fig. \ref{fig:dataset}.

\subsection{Statistics and Features}
Table \ref{tab:data_stat} shows the dataset statistics. 
For detailed video statistics such as video categories, please refer to \cite{Wang2023InternVidAL}, as the videos used in InternVid-G are an unbiased-sampled subset of InternVid-10M-FLT.

\begin{table}[t]
    \centering
    \caption{Dataset Statistics of several temporal video grounding datasets.} \label{tab:data_stat}
    \begin{tabularx}{\textwidth}{l|M{1.4cm}M{1.4cm}M{1.4cm}M{1.4cm}M{1.4cm}M{1.4cm}}
        \toprule
        & \#videos & avg. neg. span len (s) & \#queries & avg. query len (words) & avg. pos. span len (s) & video source \\
        \hline
        DiDeMo & 10642 & 29.3 & 41206 & 7.5 & 6.9 & Flickr \\
        Charades-STA & 9848 & 30.6 & 16124 & 6.2 & 8.1 & Activity \\
        ActivityNet-Captions & 14926 & 117.6 & 71957 & 14.4 & 37.1 & Activity \\
        \hline
        InternVid-G (ours) & 83614 & 203.43 & 715489 & 9.7 & 4.4 & YouTube \\
        \bottomrule
    \end{tabularx}
\end{table}

\section{HawkEye}
\subsection{Textual Representation of Video Clips} \label{sec:textual_representation}

To enable video-text LLMs to perform temporal video grounding in a text-to-text manner, another challenge is to design a representation method for LLMs to represent a segment from the video with pure text. As the visual input of video-text LLMs consists of \texttt{num\_frames} frames, an intuitive method is to tell the LLM in prompt how many frames are there in total and the timestamp of each frame. For example, ``\texttt{The video contains \%d frames sampled at \%.lf, \%.lf, ... seconds}'', where \texttt{\%d} is an integer representing the number of input frames, and \texttt{\%.lf} is a float number representing the timestamp of each frame in seconds. 
This prompt can guide LLMs to output the start and end frame number of videos like ``\texttt{From frame 3 to frame 5.}'', or the video timestamps with a format like ``\texttt{From second 4 to second 12.}'' However, in 
Sec. \ref{sec:temporal_grounding}, we show that this \textbf{frame-level} or \textbf{second-level} representation methods results in relatively poor performance, probably due to the numbers and timestamps in the prompt are difficult for LLMs to understand and analyze precisely.

Instead, we design a \textbf{coarse-grained} representation method. We categorize segments of a video into four classes: ``\textbf{beginning}'', ``\textbf{middle}'', ``\textbf{end}'' and ``\textbf{throughout}''. If the length of a video segment is larger than half the length of the entire video, we categorize this segment as ``\textbf{throughout} the entire video''. If the entire segment is in the first half of the video, we categorize this segment as ``at the \textbf{beginning} of the video''. If the entire segment is in the second half of the video, we categorize this segment as ``at the \textbf{end} of the video''. Otherwise, we categorize this segment as ``in the \textbf{middle} of the video''. An illustration of this categorization is shown in the appendix.
On the contrary, if the LLM generates a mention such as ``at the beginning of the video'' or ``throughout the entire video'', we will know that it is referring to the first half of the video or the entire video. 
This representation method only requires the LLM to provide an approximate location instead of precise start and end points of the video segment, and we design the recursive grounding to achieve the exact timestamps of the targeted video segment as output, which is described in the next section. In a word, using coarse-grained representation method can achieve the same output format as like the second-level method. However, we use the coarse-grained method as the \textbf{intermediate representation} that is easier to learn for LLMs than the very precise second-level representation, which can significantly reduce the expression difficulty for LLMs to understand.

\subsubsection{Recursive Grounding.}
However, by solely using this representation method, a model can only represent a segment with full length or half-length of the video. 
Motivated by the idea of binary search, we propose recursive grounding, which enables the model to represent shorter video segments via multiple rounds of expression. An example of recursive grounding is shown in Fig. \ref{fig:case}. Intuitively, the model first watches the entire video by sampling \texttt{num\_frames} frames and determine an approximate interval of the segment of interest. In the next round, the model focuses on the interval found in the previous round, sample another \texttt{num\_frames} frames in this interval with a higher frame rate since the length of video interval is reduced, and further narrow down the range of the interval again, serving like a video binary search.
A pseudocode of this process is presented in the appendix. Though recursive grounding may not cover all corner cases, the theoretical upper bound of its precision is still very high, and it serves as a good trade-off between precision and expression difficulty.

\subsection{Training Process}

\begin{figure}[t]
  \centering
  \includegraphics[width=0.8\textwidth]{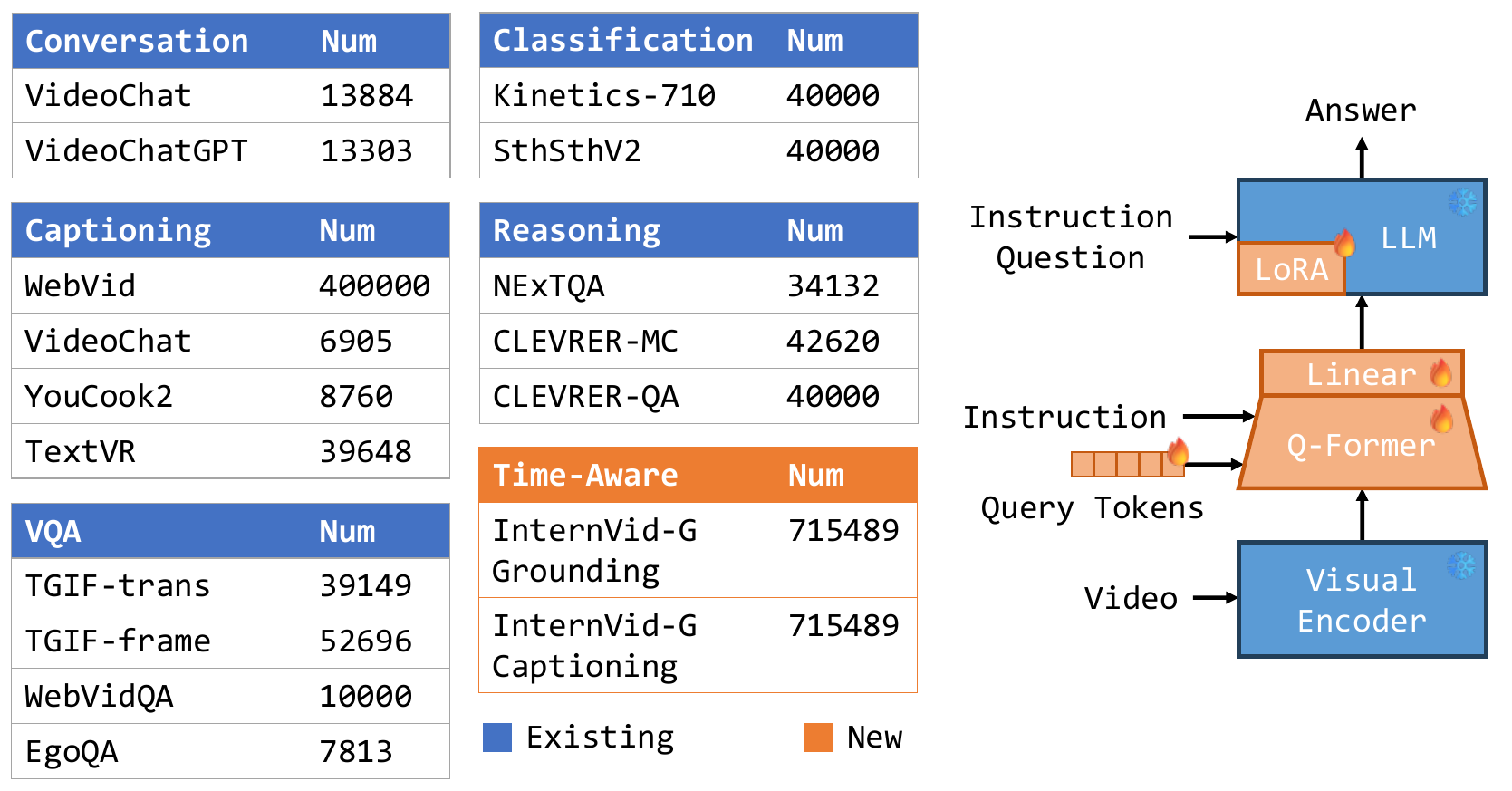}
  \caption{Datasets used and parameters trained in the instruction tuning of HawkEye.}
  \label{fig:training}
\end{figure}

The training process is shown in Fig. \ref{fig:training}.
We initialize HawkEye with the stage 2 checkpoint of VideoChat2 \cite{Li2023MVBenchAC}. We make modifications to the instruction tuning data (VideoChat2-IT) used in stage 3. Due to the limited computing resources and targetting video grounding, we only use video instruction data, and remove image data. We add two time-aware tasks based on InternVid-G to VideoChat2-IT: \textbf{temporal video grounding} and \textbf{video segment captioning}.

\begin{figure}[t]
  \centering
  \includegraphics[width=0.95\textwidth]{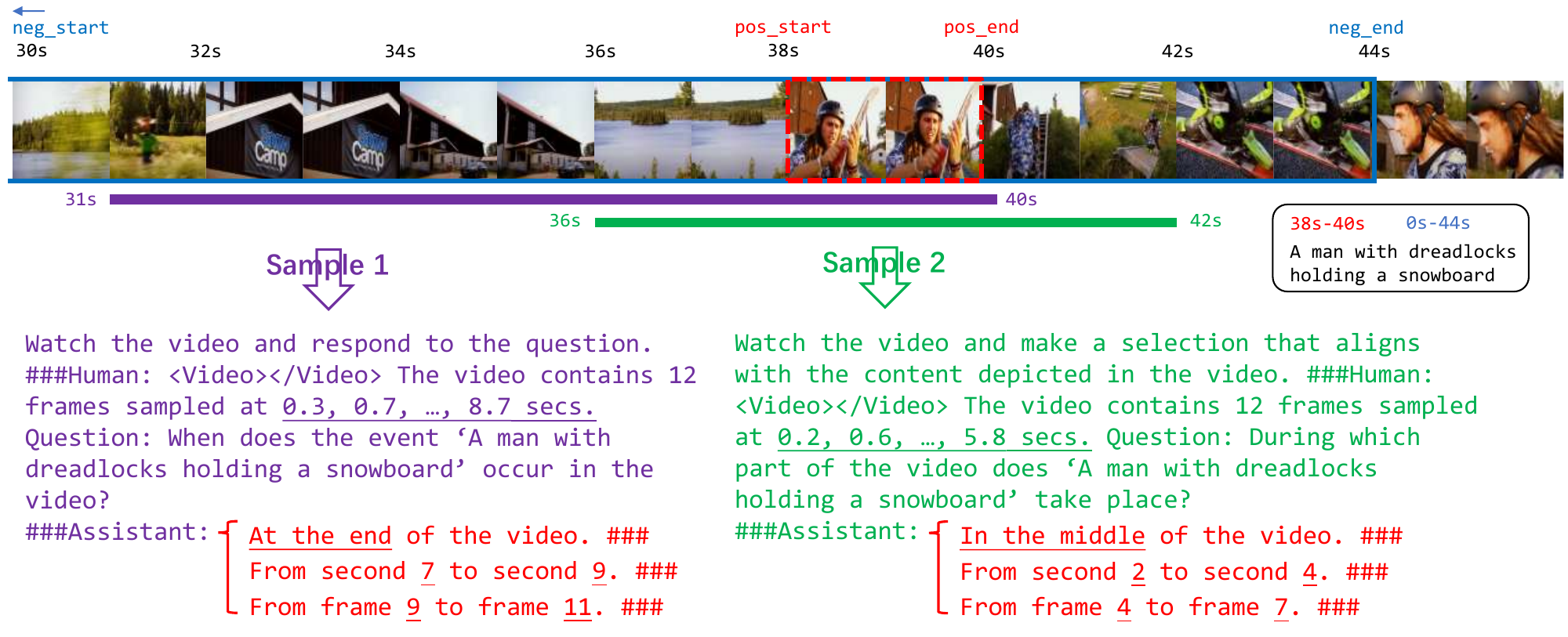}
  \caption{A demonstration of random cropping to generate the cropped video input for our model. The differences between two cropped video samples are emphasized with underlines, they share the same query and target video segment but not the same answers. The frame-level and second-level representation are also presented in \textcolor{red}{red} lines.}
  \label{fig:random_crop}
\end{figure}

Temporal video grounding is formatted as a multiple-choice question answering task. We use the query as input and ask the model to choose one of the following 4 temporal statements: ``At the beginning of the video.'', ``In the middle of the video.'', ``At the end of the video.'' and ``Throughout the entire video''. 
Importantly, we crop the video input by sampling the start position of the video in the interval of $[\texttt{neg\_start}, \texttt{pos\_start}]$ and the paired end position in the interval of $[\texttt{pos\_end}, \texttt{neg\_end}]$. Note that after which, the cropped video input includes this targeted video segment (\textit{i.e.,} $[\texttt{pos\_start}, \texttt{pos\_end}]$) regarding the text query. This cropping method enables the same text query to have different answers in different epochs, which not only augments the training data, but also prevents the model from overfitting to the shortcut relation between the query and answer. Therefore, it avoids models from neglecting the video content especially when the amount of training data is small (\textit{e.g.,} fine-tuning on small temporal grounding datasets like Charades-STA \cite{Gao2017TALLTA}). The cropping method is also designed to make the four categories of video segments have roughly the same probability to occur in each epoch. 
A demonstration of the cropping process is shown in Fig. \ref{fig:random_crop}. Random cropped sample 1 and 2 both include the targeted video segment from $38s$ to $40s$ (in \textcolor{red}{red rectangle}) but with different start and end positions ($31s$ to $40s$ in \textcolor{purple}{purple} for sample 1, $36s$ to $42s$ in \textcolor{green}{green} for sample 2) from the original video.

As the training data size of temporal grounding is significantly larger than other tasks in VideoChat2-IT, to prevent the distribution of training data from being too biased against this multiple-choice task which may hurt the model's versatility, we also add a video segment captioning task: the model is asked to generate a caption of the positive span given the cropped video clip from the negative span and the temporal statement (which is the ground truth answer in temporal video grounding task). We use the same method as temporal video grounding task to randomly crop the video input.
Unless otherwise specified, we sample 12 frames from the video as visual input. We fine-tune the Q-Former, query tokens and use LoRA \cite{Hu2021LoRALA} to fine-tune the LLM, while keep the visual encoder frozen, as shown in Fig. \ref{fig:training}. We train HawkEye for 1M steps, which takes about 7 days with 8 V100 GPUs.

\section{Experiments}
\subsection{Temporal Video Grounding} \label{sec:temporal_grounding}
We validate the temporal video grounding ability of HawkEye on two popular benchmarks: Charades-STA \cite{Gao2017TALLTA} and ActivityNet-Captions \cite{Krishna2017DenseCaptioningEI}.

\subsubsection{Comparison Between Representation Methods.}
\begin{table}[t]
    \centering
    \caption{Performance of different video segment representation methods on the testset of Charades-STA. IT denotes adding InternVid-G into stage 3 instruction tuning data, FT denotes fine-tuning on the train set of Charades-STA before testing, and RG. denotes recursive grounding. Four metrics reported are mIoU and R@IoU>0.3/0.5/0.7, where the higher the better. $^{\ast}$ fails to generate well-formatted outputs for almost half samples, so we can only take the correctly formatted ones into account.} \label{tab:comp_rep}
    \begin{tabular}{l|c|c|c}
        \toprule
        Rep. Method & IT (zero-shot) & FT & IT+FT \\
        \hline
        Frame & 24.8/40.3/23.7/8.8 & 27.5/44.7/21.2/6.3$^{\ast}$ & 48.6/73.7/54.7/26.4 \\
        Second & 20.6/35.7/14.0/2.1 & 42.3/63.0/47.0/24.6 & 47.5/69.8/53.8/29.8 \\
        Coarse (w/o RG.) & 33.2/54.1/23.8/9.1 & 42.4/71.9/37.0/13.9 & 43.1/72.7/38.2/14.4 \\
        Coarse (w/ RG.) & \textbf{33.7}/50.6/\textbf{31.4}/\textbf{14.5} & \textbf{48.2}/\textbf{72.2}/\textbf{55.8}/\textbf{27.1} & \textbf{49.3}/72.5/\textbf{58.3}/28.8 \\
        \bottomrule
    \end{tabular}
\end{table}

First, we investigate the performance of three different methods of representing video segments: coarse-grained representation (with and without recursive grounding), frame-level representation and second-level representation. Results are shown in Table \ref{tab:comp_rep}. Though the performances of all representation methods after IT+FT are comparable, coarse-grained representation method performs significantly better than other alternatives when only IT or FT is applied, which shows that our coarse-grained method is more robust in a zero-shot manner across the different distributions of videos (without FT) and much easier to be understood by the LLM especially when the amount of training data is small (without IT).

We also test the robustness of all three representation formats with different number of input frames.
It is a good feature if the number of input frames can be adjusted according to computing constraints or accuracy requirements. In this experiment we train all models using IT+FT with 12 frames, and test with 8, 12 or 16 frames from the video as input. Fig. \ref{fig:num_frames} shows that though the number of input frames and timestamps of all frames are provided in the prompt, the performance of using frame numbers or seconds to represent video segments tends to drop drastically when the number of input frames changes, probably due to the numbers and timestamps in the prompt are difficult for LLMs to understand and analyze. In contrast, our coarse-grained segment representation method is robust to different numbers of input frames.  

Overall, the above experimental results demonstrate the superiority of our coarse-grained method, and thus we decide to use the coarse-grained representation method for the following experiments of VideoChat2 and HawkEye.

\subsubsection{Zero-Shot Temporal Video Grounding.}
\begin{table}[t]
    \centering
    \caption{Zero-shot performance of temporal video grounding. Four metrics reported are mIoU and R@IoU>0.3/0.5/0.7, where the higher the better. $^{\ast}$ActivityNet-Captions is used when training TimeChat, and thus the authors did not report this performance.} \label{tab:grounding_zs}
    \begin{tabular}{l|c|c}
        \toprule
         & Charades-STA & ActivityNet-Captions \\
        \hline
        Random & 20.1/30.0/18.8/6.2 & 23.0/29.0/15.1/6.1 \\
        Choice 3 rounds Upbound & 74.8/100.0/97.0/69.2 & 71.9/91.5/84.6/68.4 \\
        \hline
        VideoChat2 & 24.6/38.0/14.3/3.8 & 27.9/40.8/27.8/9.3 \\
        VideoChat2 (our impl.) & 23.4/35.4/12.6/3.0 & 28.2/41.7/28.7/9.4 \\
        \hline
        SeViLA Localizer & 18.3/27.0/15.0/5.8 & 23.0/31.6/19.0/10.1 \\
        VTimeLLM-7B & 31.2/51.0/27.5/11.4 & 30.4/44.0/27.8/14.3 \\      
        TimeChat & - / - /32.2/13.4 & Not Applicable$^{\ast}$ \\      
        HawkEye & \textbf{33.7}/50.6/31.4/\textbf{14.5} & \textbf{32.7}/\textbf{49.1}/\textbf{29.3}/10.7 \\
        \bottomrule
    \end{tabular}
\end{table}

We compare zero-shot temporal video grounding performance of HawkEye with other video-text LLMs, as shown in Table \ref{tab:grounding_zs}.
The random baseline denotes randomly choosing a span with the average length of ground truth spans from the train set. The ``\textit{choice 3 rounds upperbound}'' baseline is the best result that running recursive grounding for 3 rounds can achieve. It is obtained by choosing the best result of $4^3=64$ possible answers. SeViLA Localizer \cite{Yu2023SelfChainedIM} is a BLIP2-based LLM that generates a float number as the score for each frame. Following their original implementation, we use 64 frames as input, and uses their method of aggregation for video moment retrieval. TimeChat \cite{Ren2023TimeChatAT} and VTimeLLM \cite{Huang2023VTimeLLMEL} are two concurrently proposed video-text LLMs that are also able to perform temporal video grounding in a text-to-text manner with using either seconds or the percentage of the video. 

For VideoChat2 and HawkEye, we use 12 frames as video input. To provide a fair comparison with HawkEye, in order to reveal the role of using InternVid-G in instruction tuning, we also re-implement VideoChat2 by reperforming the training stage 3. Our implementation only uses the video instruction data of VideoChat2-IT, and similarly we only fine-tune query tokens, Q-Formers and LoRA of LLMs. In a word, the only difference between our implemented VideoChat2 and HawkEye is the use of InternVid-G with its two newly introduced time-aware instruction tuning tasks. To evaluate VideoChat2 and HawkEye, we use recursive grounding and report results with the best maximum number of recursive rounds.

Results from Table \ref{tab:grounding_zs} show that the coarse-grained method of representing video segments with ``beginning', ``middle'', ``end'' or ``throughout'' is sufficient for temporal video grounding, especially when the precision is not required to be extremely high. For instance, the accuracy upperbound of R@IoU>0.5 on Charades-STA reaches $97.0$ after 3 recursive turns, which is already much higher than the performance of state-of-the-art temporal grounding models.

Compared to other video-text LLMs, HawkEye achieves state-of-the-art performance on zero-shot temporal video grounding, despite not trained on any human-annotated temporal video grounding data and using only 12 frames as video input (in contrast, $100$ frames for VTimeLLM, $96$ frames for TimeChat and $64$ frames for SeViLA Localizer). This shows the InternVid-G dataset as well as our proposed coarse-grained representation method can be used to enhance video-text LLMs' ability on temporal video grounding effectively and efficiently.

\subsubsection{Fine-Tuned Temporal Video Grounding.}
\begin{table}[t]
    \centering
    \caption{Fine-tuned performance of temporal video grounding. Four metrics reported are mIoU and R@IoU>0.3/0.5/0.7, where the higher the better.} \label{tab:grounding_ft}
    \begin{tabular}{l|c|c}
        \toprule
        Model & Charades-STA & ActivityNet-Captions \\
        \hline
        SOTA Specialist \cite{Moon2023QueryD,Nan2021InterventionalVG} &  - / - /57.3/32.5  & 44.2/63.2/43.8/27.1 \\
        \hline
        TimeChat & - / - /46.7/23.7 & - \\
        VideoChat2 (our impl.) & 48.3/71.8/56.4/27.7 & 38.9/55.5/34.7/17.7 \\        
        HawkEye & 49.3/72.5/58.3/28.8 & 39.1/55.9/34.7/17.9 \\      
        \bottomrule
    \end{tabular}
\end{table}

\begin{figure}[t]
  \begin{minipage}[]{0.48\linewidth}
    \centering
    \includegraphics[width=\linewidth]{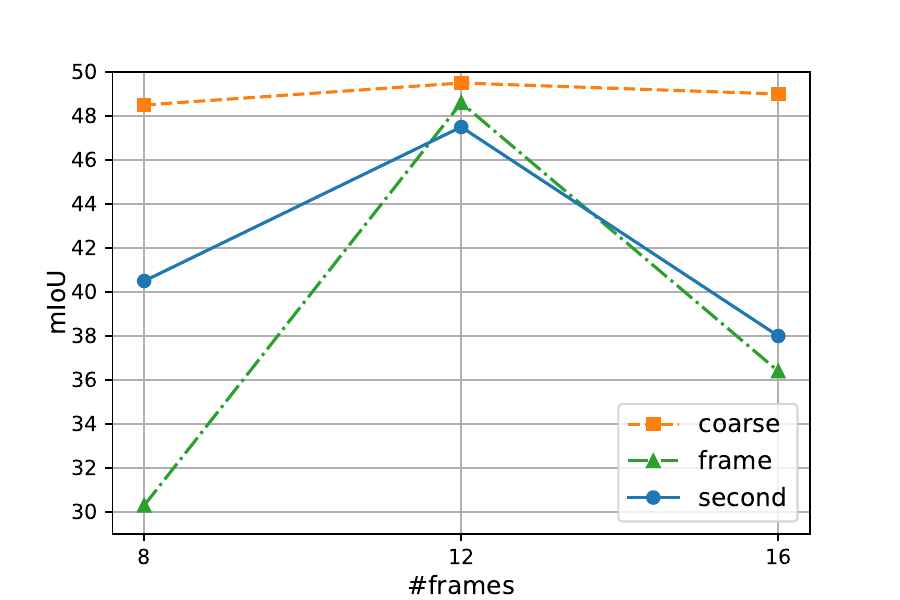}
    \caption{Performance on the test set of Charades-STA with different numbers of frames as video input during inference.}
    \label{fig:num_frames}
  \end{minipage}
  \hspace{0.02\textwidth} 
  \begin{minipage}[]{0.48\linewidth}
    \centering
    \includegraphics[width=\linewidth]{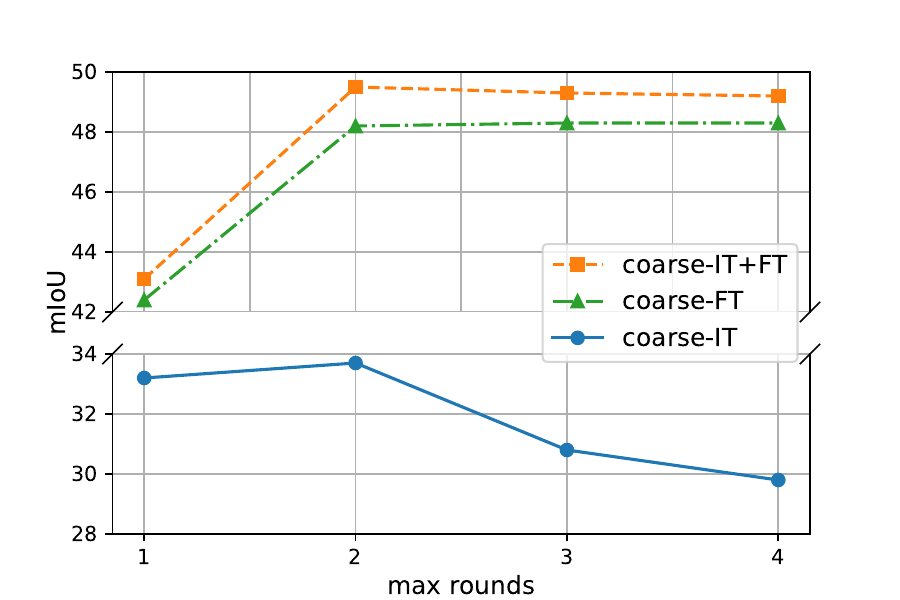}
    \caption{Performance on the test set of Charades-STA with different maximum rounds of recursive grounding.}
    \label{fig:max_rounds}
  \end{minipage}
\end{figure}

We also compare the performance of fine-tuned HawkEye with SOTA specialist models and other LLMs. 
As shown in Table \ref{tab:grounding_ft}, HawkEye outperforms VideoChat2 (our. impl.) and TimeChat, while still underperforms SOTA specialists. This indicates that though training with a large amount of time-aware samples from InternVid-G is effective, some task-specific approaches such as contrastive learning and finer-grained visual-text alignment might be required to further improve grounding performance. The results also show that our coarse-grained video segment representation method is easy for LLMs to learn and follow, as VideoChat2 without specifically instruction-tuned on time-related data have already outperformed TimeChat after fine-tuning on only thousands of examples.

\subsubsection{The Effect of Recursive Grounding Rounds.}
\begin{figure}[t]
  \centering
  \includegraphics[width=0.98\textwidth]{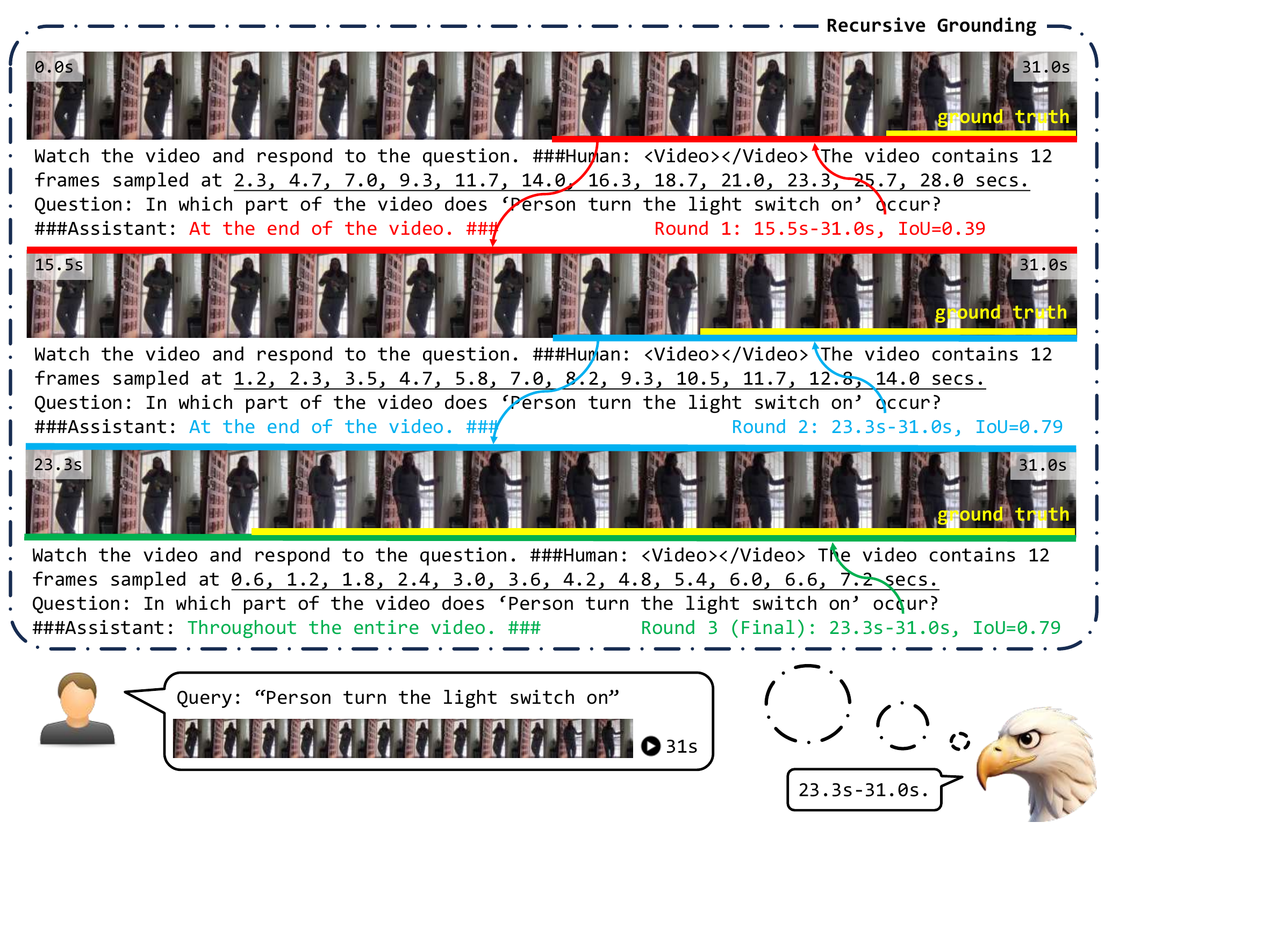}
  \caption{An example of the recursive grounding process of HawkEye. After receving the video and the query from the user, HawkEye automatically performs the grounding process with recursive grounding in the background, and converts the target span into a user-friendly \texttt{start\_sec - end\_sec} format. The four choices of the coarse-grained representation method are also described in the prompt, but here we omit them to make the presentation simpler.}
  \label{fig:case}
\end{figure}

Now we take a deeper look at how the \texttt{max\_rounds} hyper-parameter of recursive grounding affects the grounding performance, as shown in Fig. \ref{fig:max_rounds}.
When HawkEye is not fine-tuned, using smaller \texttt{max\_rounds} hyper-parameter is more likely to achieve better grounding results, as the more rounds there are, the more likely it is for the model to make mistakes on judging the location of the segment related to a query. However, after fine-tuned on the training set of the benchmark, the grounding performance hardly drops when the \texttt{max\_rounds} hyper-parameter goes larger. This is probably because the model has already gained a good understanding of the video content, so it is less likely to make mistakes and knows in which round the video is already trimmed to the targeted segment relevant to the query, and then the model can reply a ``throughout'' to break the loop. 

Comparing the performance of using only one and more than one round, we find that when only one round of localization is available (e.g., referring to a video segment in the middle of a conversation), fine-tuned HawkEye can not achieve its optimal performance. However, for the non-fine-tuned HawkEye which can be used under the majority of zero-shot conditions, this performance loss is less severe.

Fig. \ref{fig:case} shows a qualitative result of the recursive grounding process of fine-tuned HawkEye. The proposed video segment is gradually narrowed down until the ground truth takes up most of its content, in which case HawkEye generates a ``throughout'' to break the loop. 

\subsubsection{Inference Speed.} We compare the inference speed of zero-shot HawkEye and TimeChat on Charades-STA. Though a maximum of 2 rounds of recursive grounding are conducted to achieve optimal performance, HawkEye only takes an average of \textbf{1.54 seconds} for each example, while TimeChat takes an average of \textbf{1.63 seconds} as it needs to load and preprocess more frames  (96 frames for TimeChat), and its prompt is much longer to record the prcecise and miscellaneous timestamps of all the frames. This comparison confirms the efficiency of coarse-grained HawkEye even it uses recursive grounding that requires more than one round of model inference.
Refer to the appendix for more details.

\subsection{Temporal Video Grounding of Questions}
\begin{table}[t]
    \centering
    \caption{Performance of temporal video grounding of questions on the NExT-GQA.} \label{tab:nextgqa}
    \begin{tabular}{l|c}
        \toprule
         & mIoU/R@IoU>0.3/0.5 \\
        \hline
        \multicolumn{2}{l}{\textit{Fine-Tuned Models}} \\
        \hline
        Temp[CLIP] NG+ \cite{Xiao2023CanIT} & 12.1/17.5/8.9 \\
        FrozenBiLM NG+ \cite{Yang2022FrozenBilm,Xiao2023CanIT} & 9.6/13.5/6.1 \\
        \hline
        \multicolumn{2}{l}{\textit{Zero-Shot LLMs}} \\
        \hline
        SeViLA Localizer & 21.7/29.2/13.8 \\
        VideoChat2 & 24.1/36.3/16.2 \\
        VideoChat2 (our impl.) & 18.7/26.6/13.5 \\
        HawkEye & \textbf{25.7}/\textbf{37.0}/\textbf{19.5} \\
        \bottomrule
    \end{tabular}
\end{table}

In addition to retrieving relative video segments for descriptive statements, grounding questions in videos is a more difficult yet important problem, as it is a crucial step towards explainable video QA. Table \ref{tab:nextgqa} shows experiments on temporal video grounding of questions from the testset of NExT-GQA \cite{Xiao2023CanIT}, where models are required to find out a segment from the video that addresses the given question. HawkEye outperforms baseline methods with a large margin.  

\subsection{Video Question Answering}
\begin{table}[t]
    \centering
    \caption{Performance on various video question answering benchmarks without fine-tuning. $^{\ast}$: As the instruction tuning dataset VideoChat2-IT contains training data from NExT-QA, therefore this benchmark is not tested under a strict zero-shot setting for HawkEye and VideoChat2.} \label{tab:videoqa}
    \begin{tabular}{l|cccc}
        \toprule
         & MVBench & NExT-QA$^{\ast}$ & TVQA & STAR \\
        \hline
        TimeChat & 35.93 & 43.59 & 31.33 & 37.97 \\
        VideoChat2 & 51.1 & 66.5 & 40.6 & 59.0 \\
        VideoChat2 (our impl.) & 49.75 & 66.91 & 40.15 & 55.24 \\        
        HawkEye & 47.55 & \textbf{67.93} & \textbf{41.10} & 56.59 \\      
        \bottomrule
    \end{tabular}
\end{table}

As HawkEye is trained by adding two time-aware tasks from InternVid-G to stage 3 of VideoChat2, one may wonder whether introducing such a large amount of training examples from InternVid-G may lead to an unbalanced distribution of instruction tuning dataset, which will eventually impair the model's performance on other video-text tasks. 
Results in Table \ref{tab:videoqa} show that HawkEye has comparable performance with VideoChat2 (our impl., where the only difference with HawkEye is not introducing InternVid-G in instruction-tuning) on a variety of video question answering tasks \cite{Li2023MVBenchAC,Xiao2021NExTQANP,Lei2018TVQALC,Wu2021STARSR}, which shows that introducing InternVid-G in instruction tuning does not impair the model's versatility. 
In contrast, TimeChat that only focuses on time-aware tasks and wasn't instruction-tuned on diverse tasks does not have this versatility as it performs poorly on QA tasks.

\section{Conclusion}
In this paper, we explore the temporal video grounding abilities of video-text LLMs. We construct InternVid-G, a large-scale video-text corpus with segment-level captions and negative spans, which is very suitable to be used for training on temporal video grounding tasks. We propose a set of feasible practices, including constructing temporal video grounding training samples from InternVid-G, two time-aware training objectives (\textit{i.e.,} temporal video grounding and video segment captioning), a coarse-grained method of representing video segments and recursive grounding to enhance video-text LLMs' temporal video grounding abilities. Based on these findings from training HawkEye, we verify that a video-text LLM is able to perform temporal video grounding in a fully text-to-text manner. We evaluate HawkEye on a variety of temporal video grounding and video QA benchmarks to demonstrate the superiority of our method.

Future directions of this work include: (1) Introducing time-related tasks to earlier training stages such as visual-language alignment pre-training, however, which certainly requires more computing expenses.
We believe adding time-aware tasks with a large amount of training samples to the earlier training stage is very likely to improve temporal perception abilities of video-text LLMs. (2) Better representation methods of video segments. Though our coarse-grained representation method is adequate for models to express an approximate location of a video segment, a potential limitation is that it is hard to represent multiple video segments with one mention. Therefore, more representation methods are still to be explored
when we need to represent multiple video segments at a time, such as in tasks like highlight detection\cite{Lei2021QVHighlightsDM}.

%
%
\bibliographystyle{splncs04}
\bibliography{main}

\appendix
\section{Demostration of the Mapping from Video Segments to Choices}
\begin{figure}[t]
  \centering
  \includegraphics[width=0.95\textwidth]{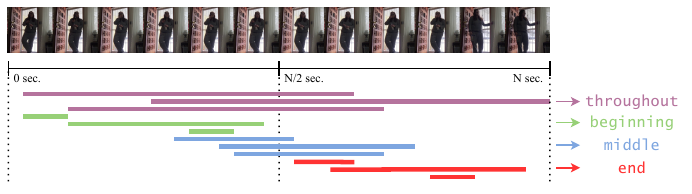}
  \caption{Illustration of the mapping from video segments to the categories of choices in our coarse-grained representation method.}
  \label{fig:representation}
\end{figure}

Fig \ref{fig:representation} shows the mapping from video segments to choices during training, which was described in Sec. \ref{sec:textual_representation}.

\section{Pseudo Code of Recursive Grounding}
The pseudocode shows the process of recursive grounding, where function \texttt{choice} denotes one round of selection. We can simply set \texttt{max\_rounds} to a large number and let the model to gradually narrow down the range of the video interval to search for more precise start and end positions, until the model thinks that almost the entire interval is the video segment it refers to and thus returns a ``throughout'' choice to break the loop. 
\begin{lstlisting}[language=Python]
def recursive_grounding(video, query, max_rounds):
    start, end = 0, get_length(video)
    for round_i in range(max_rounds):
        clip_length = end - start
        res = choice(video[start: end], query)
        if res == 'throughout':
            break
        elif res == 'beginning':
            end -= clip_length / 2
        elif res == 'end':
            start += clip_length / 2
        elif res == 'middle':
            start += clip_length / 4
            end -= clip_length / 4
    return start, end
\end{lstlisting}

\section{Training and Experiment Details}

\subsection{Prompts used in training HawkEye}
When training HawkEye on temporal video grounding task, we use the following prompt:

\noindent\mybox{gray}{\texttt{<Instruction> \newline
\#\#\#Human: <Video></Video> The video contains 12 frames sampled from \%.1f, \%.1f, \%.1f, \%.1f, \%.1f, \%.1f, \%.1f, \%.1f, \%.1f, \%.1f, \%.1f, \%.1f seconds. \newline
\#\#\#Human: Question: <Question> Options: <Options> \newline
\#\#\#Assistant: <Answer> \#\#\#}}

Where \colorbox{gray!30}{\texttt{\%.1f}} denotes the timestamp of a frame rounded to 1 decimal places, \colorbox{gray!30}{\texttt{<Instruction>}} is randomly sampled from the following 10 instructions:

\noindent\mybox{gray}{\texttt{
Evaluate the video content and select the most suitable option based on what is presented in the video.\newline
Examine the video and choose the most appropriate choice in accordance with the video's content.\newline
Watch the video and make a selection that aligns with the content depicted in the video.\newline
Analyze the video and opt for the choice that best corresponds to the content captured in the footage.\newline
Assess the video content and choose the option that aligns most closely with what is presented in the video.\newline
Evaluate the video, then select the most fitting choice based on the content portrayed.\newline
Examine the video and make a decision based on the content presented in the footage.\newline
Watch the video and choose the most fitting option based on the observed content.\newline
Assess the video content and choose a suitable option based on what is portrayed in the video.\newline
Analyze the video and select the most appropriate choice in relation to the content featured in the video.
}}

\colorbox{gray!30}{\texttt{<Question>}} is randomly sampled from the following 10 questions (\%s denotes the query):

\noindent\mybox{gray}{\texttt{
When does '\%s' happen in the video?\newline
At what time does the occurrence of '\%s' take place in the video?\newline
During which part of the video does '\%s' occur?\newline
At what point in the video does the event '\%s' happen?\newline
When in the video does the '\%s' incident occur?\newline
At which moment does '\%s' take place in the video?\newline
During which phase of the video does '\%s' happen?\newline
When does the '\%s' event occur in the video?\newline
At what time does '\%s' occur in the video sequence?\newline
When does the '\%s' situation take place in the video?
}}

\colorbox{gray!30}{\texttt{<Options>}} are the following 4 statements randomly shuffled (Of course the letters will also be modified accordingly):

\noindent\mybox{gray}{\texttt{
(A) At the beginning of the video.\newline
(B) At the middle of the video.\newline
(C) At the end of the video.\newline
(D) Throughout the entire video.
}}

and \colorbox{gray!30}{\texttt{<Answer>}} is the ground truth statement.

When training HawkEye on video segment captioning task, we use the following prompt:

\noindent\mybox{gray}{\texttt{<Instruction> \newline
\#\#\#Human: <Video></Video> The video contains 12 frames sampled from \%.1f, \%.1f, \%.1f, \%.1f, \%.1f, \%.1f, \%.1f, \%.1f, \%.1f, \%.1f, \%.1f, \%.1f seconds. \newline
\#\#\#Human: <Statement> \newline
\#\#\#Assistant: <Answer> \#\#\#}}

where \colorbox{gray!30}{\texttt{<Instruction>}} is randomly sampled from the following 10 instructions:

\noindent\mybox{gray}{\texttt{
Analyze the video content within the specified time frame and provide a detailed description of the scenes during that period.\newline
Given a specific time span, describe the activities or events taking place in the corresponding section of the video.\newline
Examine the scenes within the indicated time range and generate a textual overview of the objects, actions, and context.\newline
Provide a comprehensive narrative of the content depicted in the video during the given time span, emphasizing key elements and notable occurrences.\newline
For the specified video duration, outline the main themes and subjects present.\newline
Annotate the content within the indicated time interval, focusing on the details of people, objects, and actions captured in the video during that specific duration.\newline
Describe the visual and contextual aspects of the video scenes within the provided time range.\newline
Summarize the content of the video within the specified time span.\newline
Examine the video scenes within the given time frame and provide a detailed description of that segment.\newline
Offer a textual analysis of the video content corresponding to the specified time duration.
}}

\colorbox{gray!30}{\texttt{<Statement>}} is randomly sampled from the following 4 temporal statements:

\noindent\mybox{gray}{\texttt{
At the beginning of the video.\newline
At the middle of the video.\newline
At the end of the video.\newline
Throughout the entire video.
}}

and \colorbox{gray!30}{\texttt{<Answer>}} is the ground truth caption (text query). 

Only \colorbox{gray!30}{\texttt{<Answer> \#\#\#}} is calculated in the training loss.

\subsection{Inference Speed Experiment Details}
We first pre-process the videos into densely-extracted frames, and directly read the frames from hard disk during inference. We use AMD EPYC 7763 CPU and NVIDIA A100 GPU, and set batch size as 1. We use the default prompt of TimeChat and HawkEye respectly. Note that the speed may vary depending on the performance of hard disk, CPU and GPU, but the general conclusion is it takes about the same amount of time to run HawkEye for two rounds or TimeChat for 1 round.

\section{Case Studies}
We provide more examples of comparison between HawkEye and VideoChat2 in the appendix. Incorrect contents are marked in \textcolor{red}{red}. Hawkeye has similar abilites on answering factual questions, but performs much better on time-aware questions and understanding long-form videos.

\begin{figure}[t]
  \centering
  \includegraphics[width=0.95\textwidth]{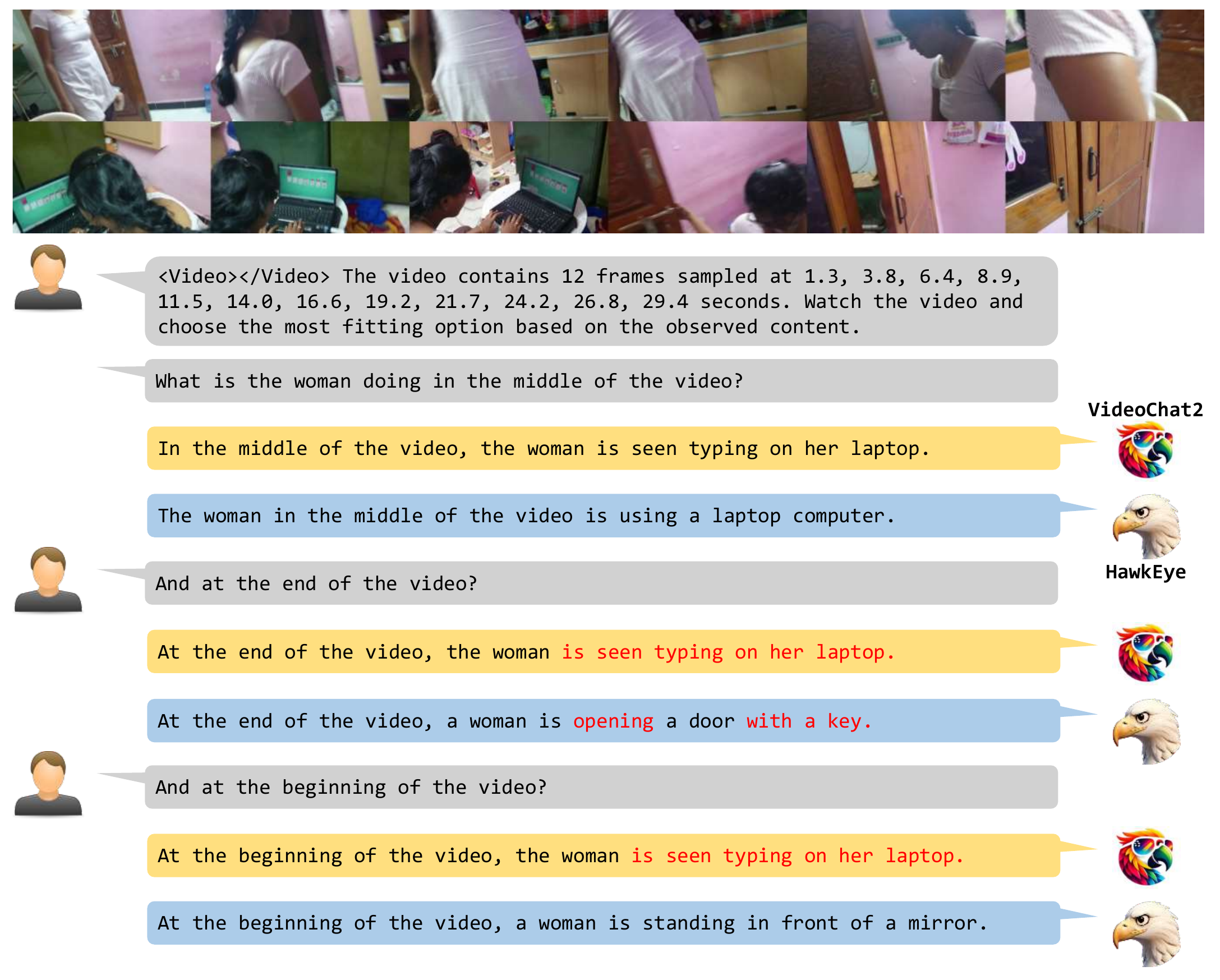}
  \caption{A Conversation with VideoChat2 and HawkEye about a daily activity video.}
  \label{fig:example1}
\end{figure}

\begin{figure}[t]
  \centering
  \includegraphics[width=0.95\textwidth]{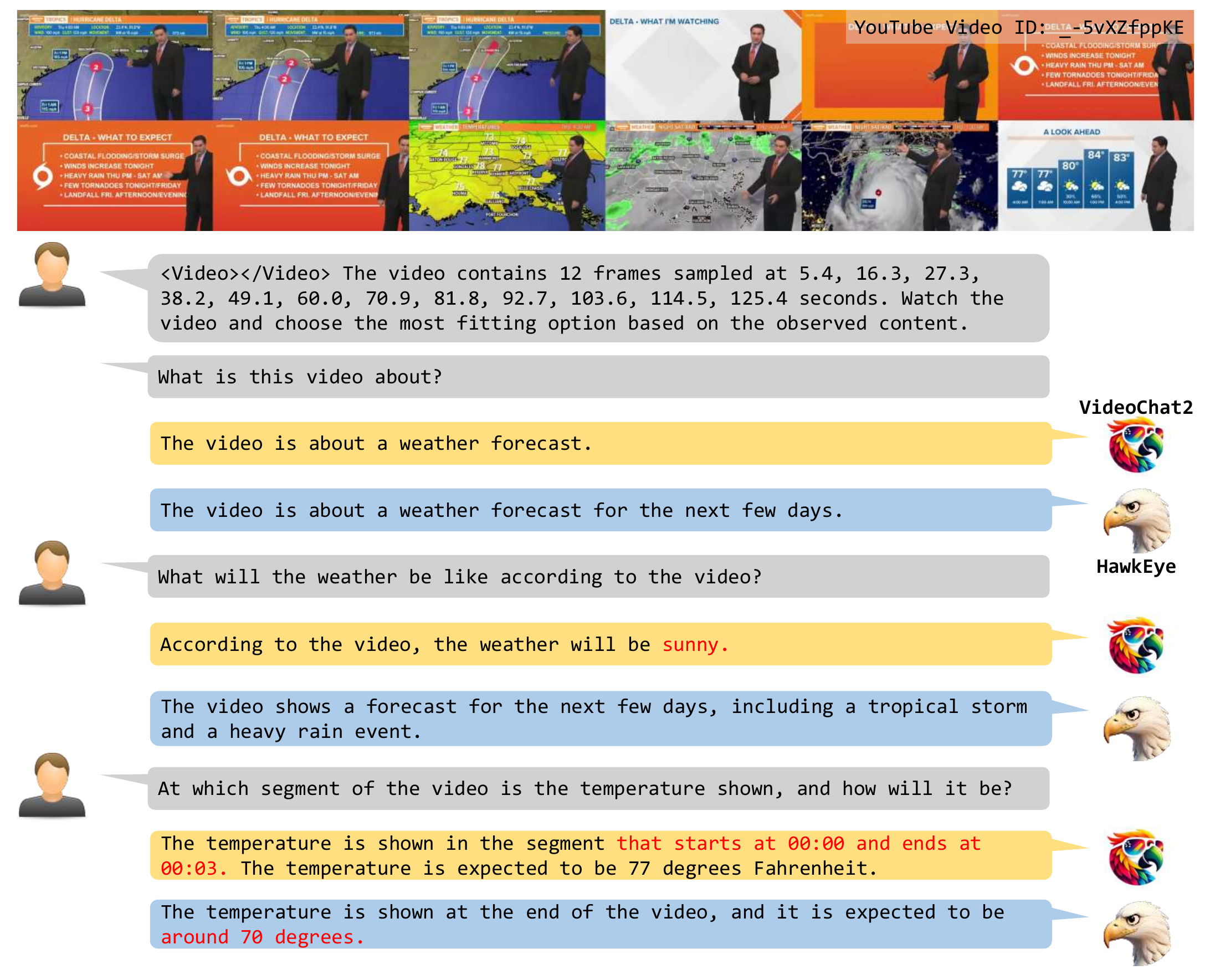}
  \caption{A Conversation with VideoChat2 and HawkEye about a news video.}
  \label{fig:example2}
\end{figure}

\begin{figure}[t]
  \centering
  \includegraphics[width=0.95\textwidth]{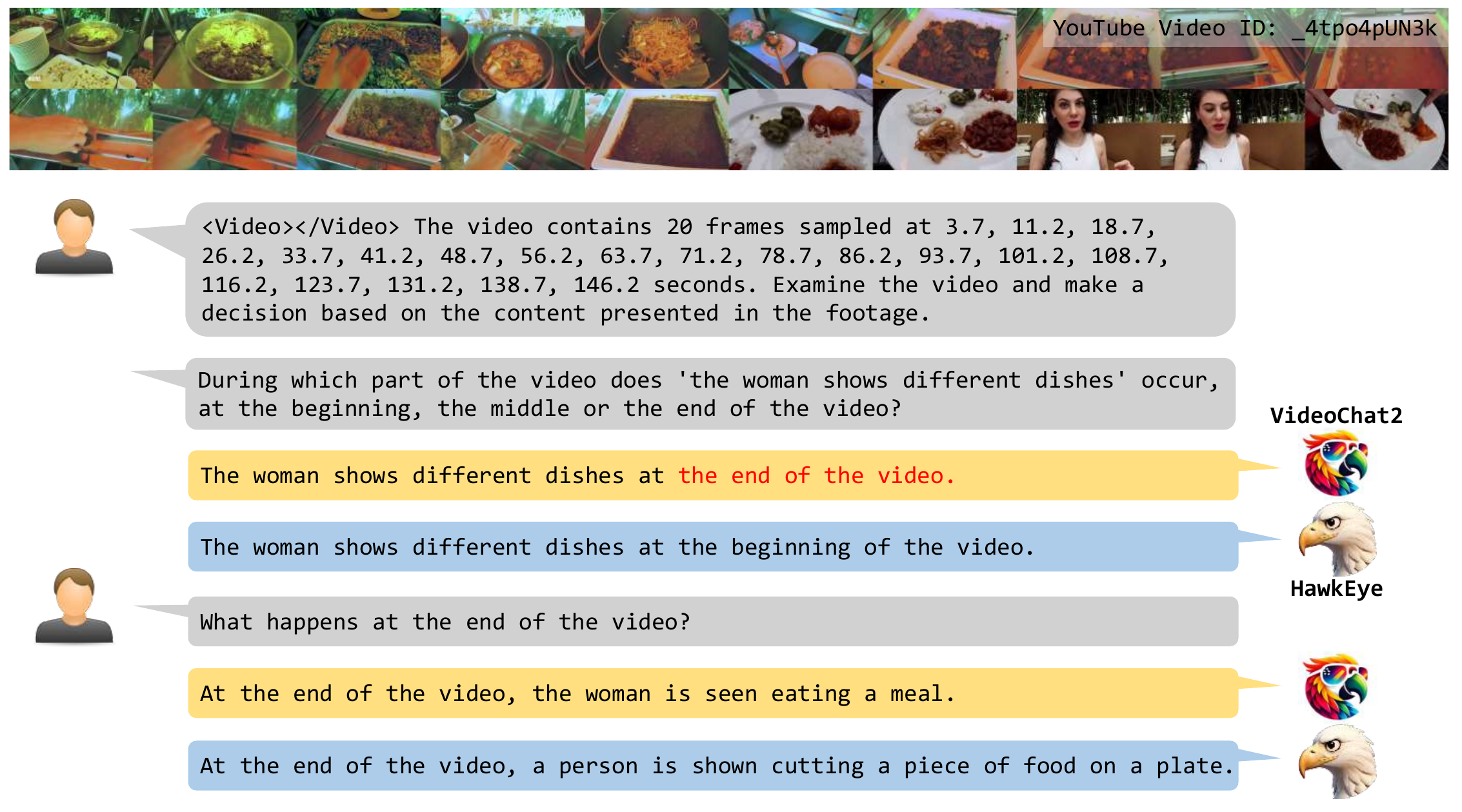}
  \caption{A Conversation with VideoChat2 and HawkEye about a vlog video.}
  \label{fig:example3}
\end{figure}

\begin{figure}[t]
  \centering
  \includegraphics[width=0.95\textwidth]{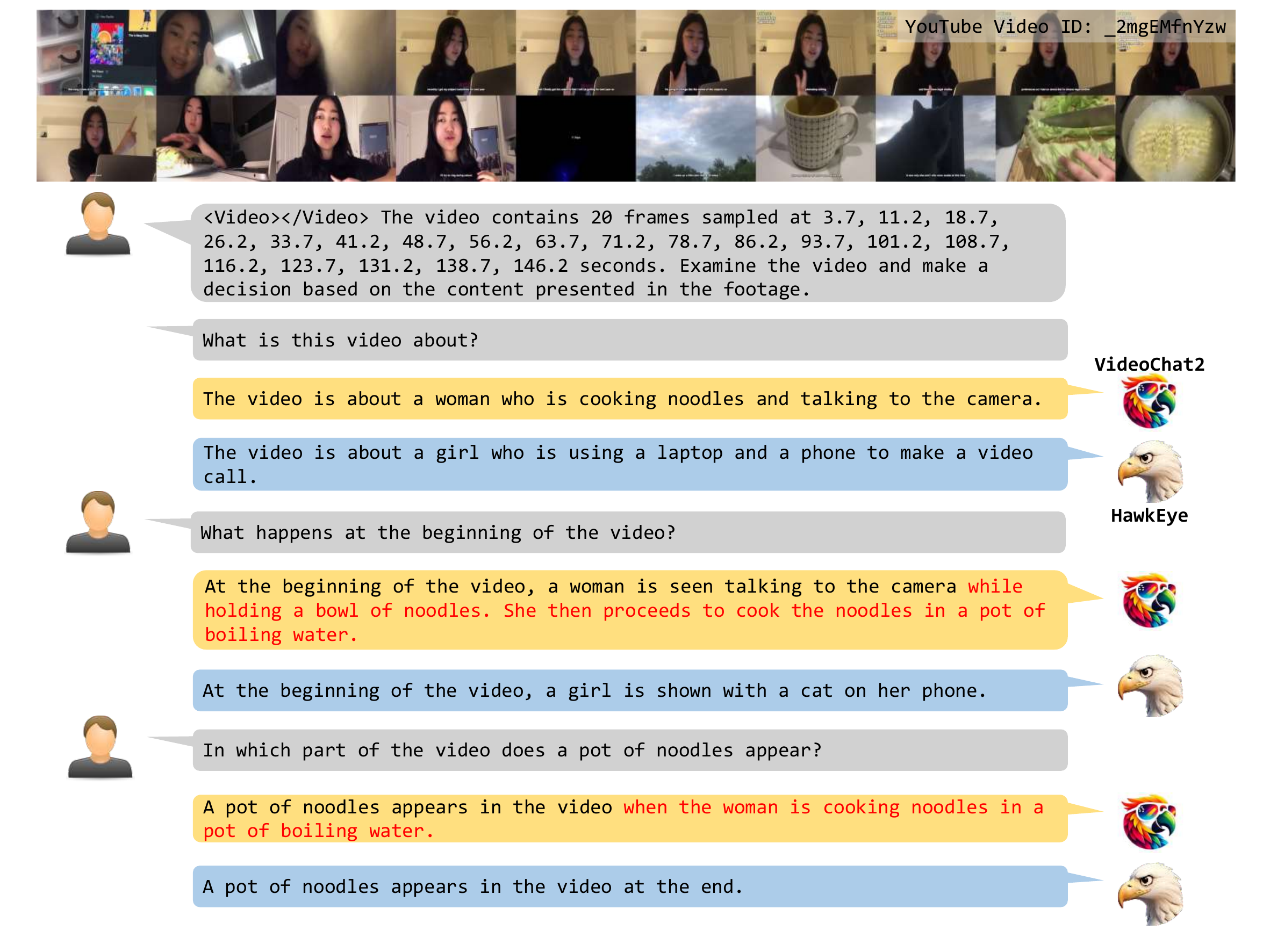}
  \caption{A Conversation with VideoChat2 and HawkEye about a vlog video.}
  \label{fig:example4}
\end{figure}

\end{document}